# Hypotheses of neural code and the information model of the neuron-detector

Yuri Parzhin

(*National Technical University "Kharkiv Polytechnic Institute" Ukraine*)

*Abstract*

*This paper deals with the problem of neural code solving. On the basis of the formulated hypotheses the information model of a neuron-detector is suggested, the detector being one of the basic elements of an artificial neural network (ANN). The paper subjects the connectionist paradigm of ANN building to criticism and suggests a new presentation paradigm for ANN building and neuroelements (NE) learning. The adequacy of the suggested model is proved by the fact that is does not contradict the modern propositions of neuropsychology and neurophysiology.*

## 1. Relevance

Throughout the history of research aimed at working out intelligent systems and artificial intelligence (AI) the choice of formalism that can be used for AI development has been discussed. The two main areas – the neural networking and semantic approach – have demonstrated their success on the one hand and have been subjected to strict criticism time and again on the other hand; this fact makes the research slack.

Some modern research [1] argued that ANN that are constructed on the basis of the known models of neural elements (NE) and use the connectionist paradigm of NE relationship establishing are just universal approximating devices that can simulate any continuous automatic machine to any given accuracy. Such ANN is rather mathematical abstraction than the model of perception and internal representation of the world made by the human brain.

As for the semantic approach, some scientific papers, e.g. [2], in particular prove that creating AI based only on representative (semantic) systems is not possible as they cannot be expressed. This property is fundamental for such systems.

Evidently, the adequate information models of neurons –NE of ANN and the approaches to ANN building should interpret and accumulate modern concepts of neuropsychology and neurophysiology about the mechanisms of information processing by neurons and brain modules.

The key point in the development of the given models can become the resolution of the problem of so-called 'neural code'. There exist a lot of approaches and a large number of scientific works devoted to this subject, but the mechanisms of coding and transformation of information in neurons and neural structures of the brain are still puzzling. One of the most interesting and fundamental studies of the neural code is the theory of vector coding suggested by E.N. Sokolov [3]. However, this theory encounters the same problems as other theories of ANN building that are based on the connectionist paradigm and use special algorithms of setting the weighting factors of NE binding to obtain the specified network response.

There is no sufficient experimental data to substantiate rigorously the information models of neurons in terms of neuropsychology and neurophysiology suggested in this paper. Therefore, we will use the hypotheses – assumptions about the neurophysiological mechanisms of certain information processes occurring in the neurons. The suggested hypotheses do not contradict the current views of neuropsychology and neurophysiology and lie on the mathematical basis, presented in [2].

## 2. Neural code hypotheses

Based on the results of research [4, 5, 6] it is possible to conclude that the morphological diversity of neurons, their functional difference depending on the position in the reflex arch point out different roles neurons play in the information processing. Thus, E.N. Sokolov, working out in his papers the detector approach suggested by D.H. Hubel, singled out specific roles of neuron-detectors, pre-detectors, neuron-modulators, command neurons, mnemonic and semantic neurons, neurons of newness and identity [3]. Experimental evidence of certain propositions of the specified approaches tells about the diversity of mechanisms (algorithms) of information processing in different types of neurons, which focus the necessity of building different information models of neurons at different steps and stages of information processing. This conclusion puts forward the following quite evident hypothesis.

*Hypothesis 1. Neurons that are different according to their morphological structure and position in the reflex arch perform different functions in information processing.*

But what information does the 'visible' neuronal response contain? What information does it react to and how does it process this information? Numerous attempts to find an answer to these questions suggest the idea that neurophysiological isomorphism of neurons response lies just in the fact of making certain sequence of spikes. The impulse activity of a particular neuron seems to tell only about its excitation to a certain degree. It seems as if the excited neuron either 'shouts' in order to outvoice other neurons or 'whispers' "*I am excited, use me for further processing*". D.H. Hubel took the same point of view; he considered that a particular neuron sends no information to other neurons except for the information about the fact of its excitation. D.O. Hebb considered the information coding as coding by the neuron ensemble [7].

Trying to answer the suggested questions, following D.H. Hubel and E.N. Sokolov, we will suggest the following statement that has neuropsycholological grounds.

***Statement 1.*** *Specific subjective perception of input image is connected with the excitation response of a single neuron or a group of specifically linked neurons. The excitation of a particular neuron is the essence of perception and the result of information processing.*

It is evident that subjective perception of the input image or its separate structural elements or characteristics (indicators of recognitions) is connected with the excitation of separate neurons-detectors. However the entire 'picture' of the internal view of a complex image, for example a 'smiling grandmother', is connected with the simultaneous excitation of a group or ensemble of neurons [5].

We suggest the following definition of the ensemble of neurons.

*A set of neurons simultaneously excited at a certain stage of information processing while perceiving or recognizing (identifying) the input image makes* **an ensemble of neurons.**

To make the models of neuronal interrelation we will specify the following definition.

The ensemble consists of neurons of different modules. Each module performs a specific function of information processing. Only one neuron can be excited at a time in each module, all other neurons in the module are either unexcited or inhibited. All the modules of the ensemble belong to the same level of information processing. The ensemble of neurons frames the initial internal presentation – 'picture' of the input image at a certain level (stage) of information processing.

*A set of neuronal ensembles at all the stages of information processing is called the* **presentation** *of the input image.*

In the given context **presentation** is not the way to submit information to someone, but the mode the brain tries to interpret internally a perceptive image.

*Information system that makes presentations which all together create a subjective picture of the world is called* **a presentative system**.

The presentative system is the first signaling system according to I.P. Pavlov, and the representative system is the second signaling system [6].

It is known that to excite a neuron at least two input information components are important – the fact of excitation of particular presynaptic neurons and the level of the excitation.

Thus, the following hypothesis can be formulated.

***Hypothesis 2.*** *The main information components of any neuron response are:*

*1) the location of the excited neuron in a certain structure (module) of the brain that is determined by its 'address' which is coded in the output signal;*

*2) the level of the neuron excitation that is coded by the frequency of generated impulses (spikes).*

Coding of the excited neuron 'address' conforms with the idea of information coding by the numbers of channels of its processing suggested by I.P. Pavlov as well as with the propositions of the vector theory suggested by E.N. Solokov.

Hypothesis 2 is theoretically grounded by the propositions of modal and vector theory of formal intelligent systems [2].

The question arises: why do the neurons send their 'address' as they are structurally interconnected by synaptic bindings?

It is necessary to take into account the fact that a neuron can receive excitatory signals from a great number of neurons, but only a part of these signals will play significant role in the process of its excitation.

But what does this 'address', which is made at the neuron output, look like?

***Hypothesis 3.*** *The "address" of a neuron in the module of information processing is coded by a unique for this module set and relation of neurotransmitters generated in the neuron.*

This hypothesis is not evident as the modern research deal with neurotransmitters only when studying the mechanism of transmitting signals of excitations or inhibition. However, the fact that there exists a great variety of neurotransmitters supports this hypothesis. Nowadays there are nearly 100 kinds of neurotransmitters [8], but the significance of this variety has not been studied yet. Besides, we know the following facts that can be explained within this hypothesis [8, 9]:

1. In one neuron more than one transmitter can be generated and released into presynaptic endings. Each neuron generates a dominating transmitter as well as a group of transmitters that act like modulators. Co-mediators and neuropeptides are generated as well.

2. A set of transmitters for a particular type of neurons is constant.

3. Transmitters are placed non-diffusely along the brain tissue but very locally in the limited centers (modules) and tracts.

It is also known that the character of synaptic action is determined not only by the chemical nature of a transmitter but by the nature of receptors of a postsynaptic cell. One postsynaptic neuron can have more than one type of receptors for a certain transmitter. Thus, the 'transmitter-receptor' binding acts like 'key-lock' relation while unlocking a set of neuronal 'gates' or ion channels of its membrane, which leads to the neuron excitation in the end.

According to the suggested hypotheses 'address' response of the excited neuron-detector is the basic information which is fed to subsequent neurons in the reflex arch, and the frequency response in this case plays a significant role in the process of 'competitive activity' of simultaneously excited neuron-detectors defending the right to participate in the further information processing.

Regarding coding the level of neuron exciting by the frequency of its impulse response, it is known that the frequency of spikes and burst activity of the neuron (neuron response pattern) depend both on the amount of excitation threshold (ET) of neuron and the amount of its membrane potential (MP) exceeding ET and on other characteristics of the neuron, for example, the refractory period. The frequency generation of spikes or neuron lability lies within the limits from 100 to 1000 Hz. MP, in its turn, in many cases is determined by summing excitatory (with '+' sign) and inhibitory (with '-' sign) postsynaptic potentials (EPSP and IPSP). The amplitude of the local EPSP is likely to depend on the amount of transmitter which is released by a presynaptic cell, i.e. on the level of its excitation. The frequency of impulses is proportional to the excess of ET by the sum of ESPS. The increase of impulsation lasts until the satiation comes – 100 % operating cycle.

The frequency coding is more apparent in sensor, afferent and efferent systems [10]. Sensor and afferent systems are featured by non-linear dependence of neuron response on the intensity of an irritant (Weber-Fechner's and Stevens' laws). However, such dependence is not evident in neurons of cortical and subcortical brain structures [11]. This fact enabled the hypothesis of coding information by the answer pattern to be put forward [12]; but this hypothesis was not confirmed definitely and was criticized. Some researchers consider the variability of the neuron frequency response to the same stimulus as the evidence of its probabilistic nature [13], as well as an incentive to putting forward new hypotheses of neuron code and new approaches to interpreting informative characteristics of neuron impulsation.

Nevertheless, Hubel's statement that 'the exact location of each impulse in a train has no importance in most cases, but their average number within a time interval is significant [5]' seems the most precise.

Experiments confirm that the electric characteristics of different neurons vary in certain limits. Thus, for example, the membrane resting potential (RP) varies within the limits from -55mV to -100mV. For motoneurons RP is on the average about -65mV, ET is nearly -45mV, local EPSP is from +0,5 to +1,0 mV, and EPSP total value that shifts MP is about +20mV. Thus, if from 20 to 40 presynaptic neurons simultaneously (spatial summation) or within a short period of time are exposed to excitation which will cause local EPSP with the amplitude of +1,0 - +0,5 mV (zero IPSP), then ET will be got over and action potential (AP) will occur. However, a real multipolar neuron has significantly greater number of relations and only certain presynaptic neurons send signals able to excite it to get over ET. This, in fact, substantiates the role of weighting coefficients of synaptic relations used in classic connectionist ANN. Besides, ET of any neuron is known to be able to change under certain conditions. Thus, when the number of excitatory synaptic inputs decrease, ET also decreases in the course of time, which means that significantly less depolarization current is necessary to cause AP and vise versa [14, 15].

Taking into account the information stated above, the following hypothesis can be suggested.

***Hypothesis 4.***

*а). A set of EPSP, the total value of which, enables getting over a neuron ET is a set of necessary and sufficient conditions for its excitation.*

*б). As local EPSP is connected to the excitatory level of specific presynaptic neuron, to identify whether this particular EPSP belongs to the set of necessary and sufficient conditions of postsynaptic neuron excitation is possible on the basis of the excitatory neuron address.*

*в). A set of 'addresses' of presynaptic neurons, the excitation of which creates a set of necessary and sufficient conditions for exciting this specified postsynaptic neuron, is made in the process of neuron learning. In the process of learning the*

*neuron memorizes only those 'addresses' of presynaptic neurons, the excitation of which with the specified excitatory level (the frequency of excitatory impulses ) will become a necessary and sufficient condition for its own excitation.*

We will state the following important definition.

*A set of 'addresses' of presynaptic neurons, the excitation of which is a necessary and sufficient condition for exciting a postsynaptic neuron bound to it, is called the* **concept** *(Con) of this neuron.*

Thus, to determine whether the 'addresses' of presynaptic neurons belong to Con of a postsynaptic neuron is possible while learning either with or without the 'teacher'. 'Teachers' are the corresponding neurons of other subsystems, for example, a representative subsystem (RS). The importance and the role of RS are considered in the paper [2]. Evidently, the more often the 'address' of a particular presynaptic neuron appears in the process of learning in the vector of input signals, the greater the probability that this 'address' belongs to its Con.

This hypothesis absolutely agrees with Hebb's classical postulate that 'if the axon of A cell is located close enough to B cell to excite it and permanently participates in its activation, then either one or both cells have such metabolic changes or growth phenomena that the efficiency of A as a cells that activates B increases '[7].

In his paper [16] Simon Haykin cites one of the modern interpretations of Hebb's postulate modified for connectionist ANN learning, where the postulate includes two rules of changing weighing coefficients of synaptic relations of neurons – NE of ANN:

1. If two neurons along the both sides of synapse (binding) are activated simultaneously (synchronously), the binding strength increases;

2. If two neurons on both sides of synapse are activated asynchronously, then the synapse goes down or even dies off.

We can see that this interpretation contains rule 2, which is absent in the classical statement. Actually, in some types of neurons there is the process of competitive decrease of the number of synaptic bindings. The vanish of polyneuron innervation of motoneurons during the period of nervous system development can be an example of such competition. But if a real 'mature' neuron has synaptic binding with another neuron, then it is evidently important; and to destroy or degrade such binding, the periodic change of presynaptic neuron activity is not sufficient [8].

But formulating learning rules relying on the results of external display of cause-and-effect relation of presynaptic effect and postsynaptic cell response, that are shown in Hebb's postulate, is quite controversial, as these rules are rather superficial. Thus, according to their content they should comprise self-learning, but they say nothing about learning steps, learning time, statistical character of synapse weight change, the requirements to the learning sequence of input signals. The formal use of these rules demands constant feedback of postsynaptic and presynaptic neurons and can cause at least three negative consequences:

- the change of synapse weights after each learning step; and while self-learning this step can be determined by the any vector of input signals;

- infinite increase of weights of synapses that are permanently or periodically activated during limitless learning (if the saturation threshold is established, then all periodically activated synapses will reach it early or late);
- degradation of significant binding when the vector of input signals is incomplete or 'false'.

This approach to neuron leaning is in no way connected with the processes of information processing by the neuron, the neuron acting like a 'blackbox'.

## 3. Information model of neuron-detector

Grounding on the suggested hypotheses we can make an information model of neuron detector.

Let detector $d$ has $n$ number of synaptic inputs and one output. For simplification we assume that the detector does not have inhibitory input synapses. The synaptic input $i$ can receive excitatory signal $x_i(a,b)$ from presynaptic neuron with the 'address' $a$ and the level of excitation $b$.

When detector $d$ is excited, the normalized signal $y'(a',b')$ is generated in its output, where $a'$ is its address component, $b'$ is the level of excitation. Detector $d$ also has control input $z$, where the control signal from the neurons of other systems, e.g. from RS, is fed.

The detector has the following characteristics:
- excitation threshold – $et$;
- resting potential – $rp$;
- shift in the membrane potential – $smp$;
- action potential – $ap$.

We introduce $w_i$ function of membership of signal $x_i$, that is fed to $i^{th}$ input of detector $d$, on its concept $Con(d)$.

For learning 'with teacher' $w_i$ looks like (1):

$$w_i = \frac{l_i(t_0)}{k(t_0)} \quad (1),$$

where: - $l_i(t_0)$ is the total number of signals $x_i$ fed to a $i^{th}$ input of detector $d$ during learning time $t_0$. During learning time $x_i$ is fed synchronously with $z$;

- $k(t_0)$ is the general amount of cycles of feed of input signals $\overline{X}$ vectors to inputs $d$ at the moments of learning time $t_0$.

Then,
$$\text{if} \quad \begin{cases} w_i = 1, \text{ then } x_i(a) \in Con(d); \\ w_i < 1, \text{ then } x_i(a) \notin Con(d). \end{cases} \quad (2)$$

Probably, detectors always learn 'with teacher', neurons of different brain systems acting like teachers.

However, the option of detector self-learning can be formalized. In this case $w_i'$ will look like a power function:

$$w_i' = \left(\frac{l_i}{k}\right)^c \qquad (3),$$

where $0 < c < 1$.

In this case to determine the membership of $x_i$ on concept $Con(d)$ it is necessary to introduce the statistical threshold of membership $q < 1$ (Fig.1).

Then,

if $\begin{cases} w_i' < q, \text{ then } w_i = 0 \text{ and } x_i(a) \notin Con(d); \\ w_i' \geq q, \text{ then } w_i = 1 \text{ and } x_i(a) \in Con(d). \end{cases}$ (4)

When input excitatory signals are received, *smp* of detector *d* is:

$$smp(d) = \sum x_i(b) | x_i(a) \in Con(d) \qquad (5).$$

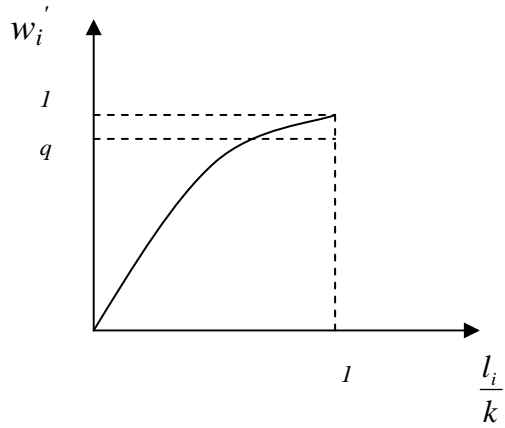

Fig.1 Power function $w_i'$

Then, if $rp(d) + smp(d) \geq et(d)$, then there exists $ap(d) = y(b')$, where $y(b')$ is non-normalized component of output signal – non-normalized level of excitation of detector *d*.

In the suggested information models of NE it is convenient to use NE positive characteristics. Besides, to create ANN on the basis of these models it is reasonable to use NE interrelations in the module of information processing of 'general line' type rather than radial structure of interrelations between pre- and postsynaptic neurons.

Then $y(a',b')$ will exist under the following condition:

$$(g_0 + \sum x_i(b) | x_i(a) \in Con(d)) > g^* \qquad (6),$$

where: - $g_0$ is the basic value similar to $rp(d)$;
- $g^*$ is the threshold value similar to $et(d)$.

Values $g_0$ and $g^*$ set the 'corridor' of excitation determining the length of $Con(d)$ and the level of excitation of input signals that belong to $Con(d)$.

Value $x_i(b)$ in the formula (6) is variable and can change within the limits:

$$x_i(b)_{min} < x_i(b)_{opt} < x_i(b)_{max} \quad (7).$$

Values $x_i(b)$, fixed at the moment of learning and statistically can be found more often in the input vector of signals will be the most optimal values $x_i(b)_{opt}$, and these are the values to determine values $g_0$ and $g^*$. In this case, value $g_0$ assumes a character of supporting value for getting over $g^*$ under values $x_i(b)$ ($x_i(a) \in Con(d)$) that are less than optimal.

Value $y(b')$ under condition (6) will be determined by the expression:

$$y(b') = g_0 + g' + g'' \quad (8),$$

where
$$\begin{cases} g' = \sum x_i(b)/x_i(a) \in Con(d) \\ g'' = \sum x_i(b)/x_i(a) \notin Con(d); \end{cases} \quad (9);$$

- $g'$ is the value that determines the contribution of input vector signals that belong to $Con(d)$ to the total EPSP;
- $g''$ is the value that determines the contribution of input vector signals that do not belong to $Con(d)$ to the total EPSP.

$$y(b') = \Delta g + g'' \quad (10);$$

where $\Delta g$ is the value necessary to get over the excitation threshold $g^*$.

$$\Delta g = g_0 + g' \quad (11).$$

In the process of competition of simultaneously excited detectors value $y(b')$ is important to determine one leader in a module, signals from which will participate in the further information processing.

*We call this process **α- competition** [2].*

Actually, there cannot exist two similar concepts, but it is possible that $Con(d_1) \subset Con(d_2)$. Then detectors $d_1$ and $d_2$ will be excited simultaneously, but value $y_1(b')$ of detector $d_1$ will be less than value $y_2(b')$ of detector $d_2$.

In the context of α-competition output signals $y_i(b')$ of simultaneously excited NE-detectors are compared in comparators $C$ of each detector, in case if any external signal $y_i(b')$ exceeds its own one, an internal control signal of inhibition $h$ is generated. This signal 'drops' the excitation of detector switching it over into the state of 'pre-excitation'. Fig. 2 shows the diagram of generating $h$ signal in $d_1$ detector under the condition of $y_2(b') > y_1(b')$.

In case if $y_1(b') > y_2(b')$, the detector $d_2$ is inhibited. Then $C$ comparator of detector $d_1$ sends $y_1(b')$ signal further to unit $N$, where function $f(y_1(b')) = y_1'(b')$ is performed. This is the function of normalization of excitation level $y_1(b')$ and generation of output signal $y_1'(a',b')$ with the normalized component of excitation level $y_1'(b')$ and address component $y_1'(a')$ (Fig. 2).

Perhaps for real neurons, the transmission of signals $y_i(b')$ in a neuron can be performed through glial cells using electric synapse or through axoaxonic synapses.

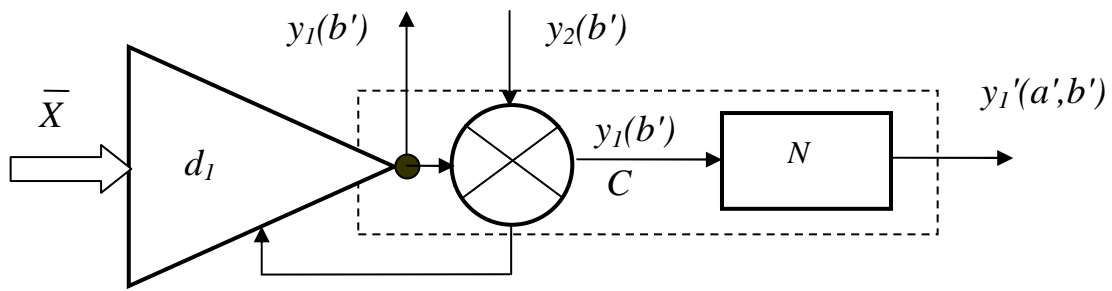

Fig. 2. Diagram of forming $h$ and $y_1'(a',b')$ signals

The normalization of the detector excitation level in output signal $y_i'(b')$ is necessary to prevent the effect of its limitless increase in NE cycle of ANN, which corresponds to real processes of normalization (saturation) of an output neuron signal.

Fig.3 shows the diagram of forming output signal $y_i(b')$ by NE-detector $d_i$ when the vector of signals $\overline{X}$ is fed to the inputs.

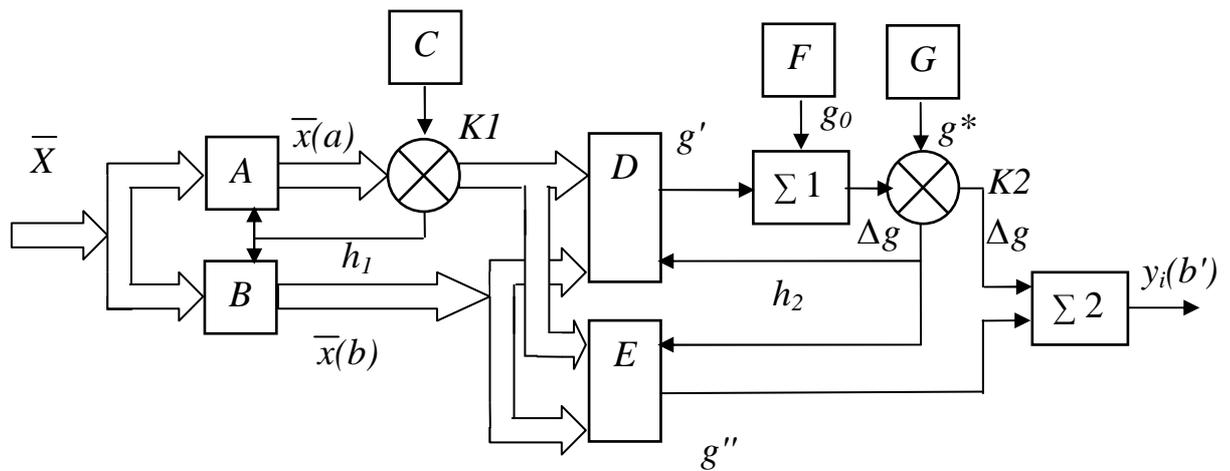

Fig. 3 Diagram of forming output signal $y_i(b')$

This diagram includes the following elements:
- $\overline{X}$ is the input vector of signals;
- *A* is the unit of determination of the addresses of presynaptic NE and formation of input vector $\overline{x}(a)$;
- *B* is the unit of determination of the levels of excitation of presynaptic NE and formation of input vector $\overline{x}(b)$;
- *C* is the memory unit of $w_i$ membership functions of input signals $x_i$ on concept *Con(d)* of NE-detector;

- *K1* is the comparator implementing the membership of input vector $\bar{x}(a)$ on concept *Con(d)* of NE-detector;
- $h_1$ is an inhibitory signal formed in *K1* in case of absolute discrepancy between input vector $\bar{x}(a)$ and concept *Con(d)* of NE-detector. Signal $h_1$ 'drops' values *A* and *B*;
- *D* is the unit of forming $g' = \sum x_i(b)/x_i(a) \in Con(d)$;
- *E* is the unit of forming $g'' = \sum x_i(b)/x_i(a) \notin Con(d)$;
- $\sum 1$ is the adder calculating value $\Delta g = g_0 + g'$;
- *F* is the unit of memory of $g_0$;
- *G* is the unit of memory of $g^*$ – excitation threshold of NE-detector;
- *K2* is the comparator setting the exceeding of value $\Delta g$ of quantity $g^*$. In case when $\Delta g < g^*$, the inhibitory signal $h_2$, which 'drops' values $g'$ and $g''$ is generated. Otherwise *K2* sends value $\Delta g$ to the input of the adder for $\sum 2$ to from signal $y_i(b')$.

The process of NE-detector learning 'with teacher' is shown in Fig. 4, where the following elements can be seen:

- *A* is the unit of determination of the addresses of presynaptic NE and formation of input vector $\bar{x}(a)$ if the following condition $\bar{x}(a) \cap Con(d)$ is met;
- *B* is the unit of determination of the levels of excitation of presynaptic NE and formation of input vector $\bar{x}(b)$ if the following condition $\bar{x}(a) \cap Con(d)$ is met;
- *C* is the unit of counters $l_i(t_0)$;
- *D* is the unit of forming control learning signals $z'$ when input signal $z$ is received;
- *E* is the counter $k(t_0)$;
- *F* is the unit of calculating $w_i$ membership functions of input signals $x_i$ on concept *Con(d)* of NE-detector;
- *G* is the unit of memory of $w_i$ membership functions of input signals $x_i$ on concept *Con(d)* of NE-detector;
- *H* is the unit of memory of optimal values $x_i(b)$ when $x_i(a) \in Con(d)$;
- *K1* is the comparator comparing values $x_i(b)$ received at the moment of time $t_0$ (under condition $x_i(a) \in Con(d)$) with optimal values of these quantities;
- *J* is the corrector of optimal values $x_i(b)$;
- *I* is the unit calculating $g^*$;
- *K3* is the comparator defining the value of displacement of newly formed value $g^*$ and the value stored in the memory unit *M*;
- *L* is the corrector of $g^*$;
- *K2* is the comparator comparing values $x_i(b)$ received at the moment of time $t_0$ (under condition $x_i(a) \in Con(d)$) with minimum values of these quantities;
- *N* is the unit of memory of minimum values $x_i(b)$ when $x_i(a) \in Con(d)$;
- *Q* is the corrector of minimum values $x_i(b)$;
- *T* is the corrector of base value $g_0$;
- *P* is the unit of memory of value $g_0$.

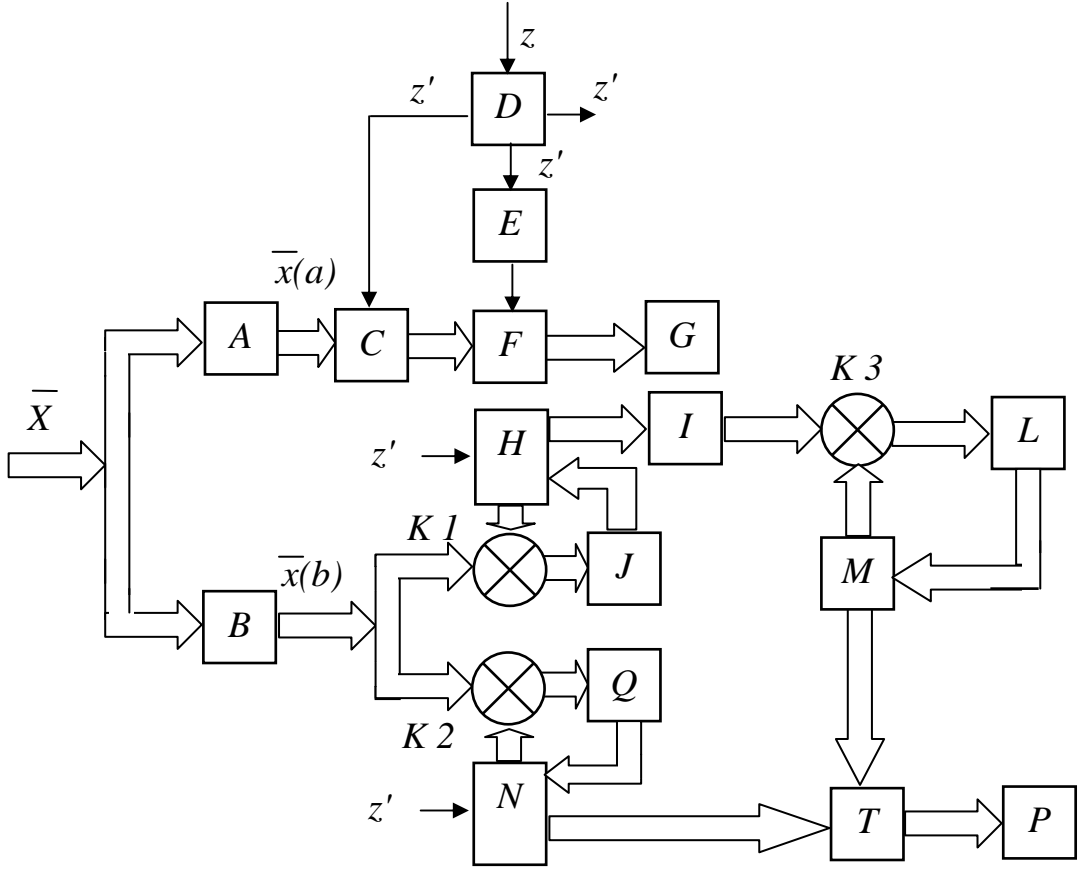

Fig. 4. Diagram of NE-detector learning 'with teacher'

Control learning signal $z$ is formed by NE-detector of other systems, for example, representative system. We will study the process of NE-detector of presentative system (PS) learning under the control of NE-detector of representative system (RS). To do this PS detector should establish relation with the corresponding RS detector. The diagram of establishing this interrelation is shown in Fig. 5.

We assume that at the initial moment of time detector $d_i$ is not activated and is a 'free' state. This state is characterized by the fact that when any vector of $\overline{X}$ signals comes into its inputs it is not excited, since $Con(d_i)$ is not formed. The detector also does not recognize the learning signals $z$ since $d_i$ is not bound to RS learning detector $d_i^*$. This binding is formed as a result of memorization of 'address' $d_i^*$ by detector $d_i$.

To activate $d_i$ detector it must be 'captured'. To do this command neuron $H$ that controls the process of activation of all the neurons of the module is required. All output signals of NE module and input vector of $\overline{X}$ signals come into $H$ inputs. If $\overline{X}$ comes into the module, but no NE of the module is excited, then signal 'capturing' free NE is generated at the output of $H$. Therefore, command neuron $H$ is the neuron of newness. When free detector $d_i$ nearest to $H$ receives control signal $v$, it interprets input vector $\overline{X}$ as the vector of identification and memorizes it as initial $Con(d_i)$. Thus, $d_i$ becomes detector-identifier of $\overline{X}$ sample. In case if there is no further learning 'with teacher' of $d_i$ detector, it will be excited only when vector $\overline{X}$ comes into its inputs or it will learn on its own.

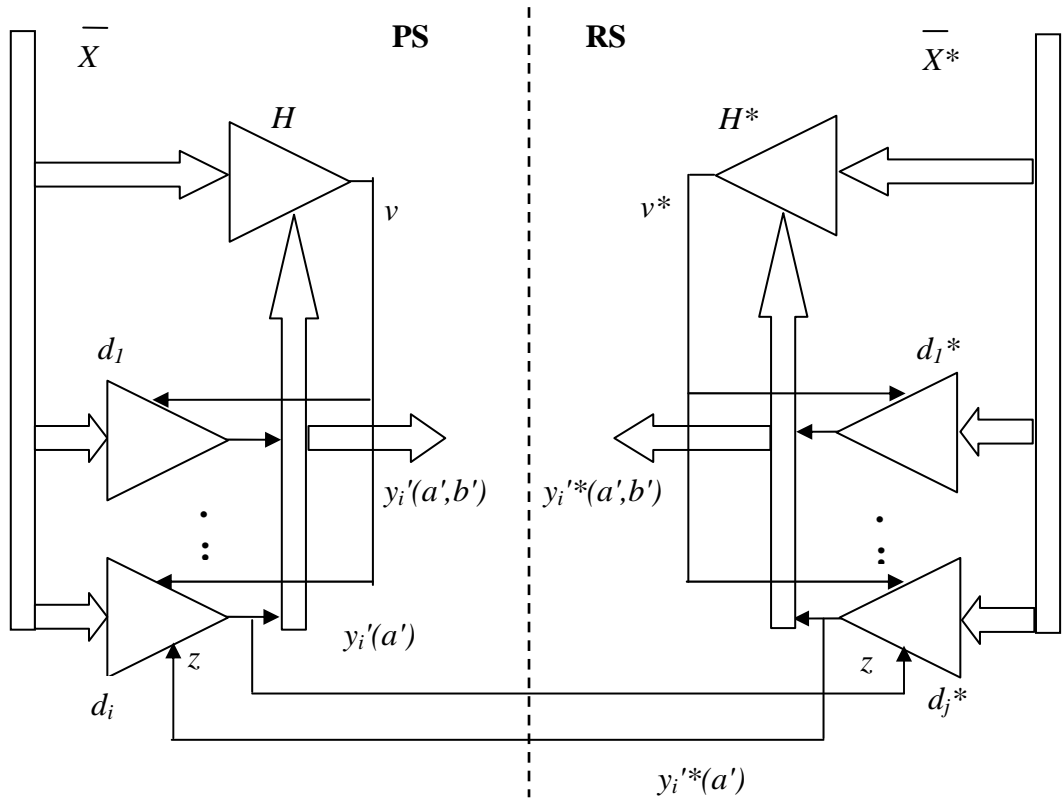

Fig. 5. Diagram of establishing interrelation between $d_i$ and $d_i^*$ detectors

We assume that $d_i$ is activated, but its interrelation with $d_i^*$ learning detector is not established yet. Then, $\overline{X}$ vector comes into $d_i$ input it is excited and sends $y_i'(a')$ signal to the module of RS. If $d_i^*$ detector is excited simultaneously with $d_i$ in RS module, then $d_i$ detector interprets its output signal $y_i'^*(a')$ as control learning signal $z$. In this case $d_i$ detector memorizes the value of signal $y_i'^*(a')$ and thereby the interrelation between $d_i$ and $d_i^*$ is established. Similar process occurs in detector $d_i^*$ when it memorizes output signal $y_i'(a')$ of detector $d_i$. Thus, detectors $d_i$ and $d_i^*$ learn from each other.

When vector $\overline{X}' \cap Con(d_i)$ comes into $d_i$ input and if there exists control learning signal $z$, learning $d_i$ takes place as a result of correction $Con(d_i)$.

If in $z$ existence vector $\overline{X}'$, which does not intersect $Con(d_i)$, comes into $d_i$ inputs, then $d_i$ detector is nor excited. In this case 'capture' takes place as well as excitation of a new detector with the alternative for this signal $z$ concept.

If vector $\overline{X}' \cap Con(d_i)$ comes into $d_i$, inputs, but learning signal $z$ that occurs simultaneously does not coincide with the memorized value $y_i'^*(a')$, then $d_i$ detector is not excited (it is inhibited). Then 'capture' takes place and excitation of new detector $d'_i$ with $Con(d_i')=\overline{X}'$ by new value of $z$ signal.

If input vector does not come into $d_i$ inputs, but real $z$ signal comes, then *associative excitation* of this detector takes place. The mechanism of this kind of excitation is not considered in this paper.

## 4. Conclusion

NE-detectors are ANN elements that converge information flows. If ANN comprised only NE-detectors suggested in the paper, it would be little different from classical connectionist ANN. But real neuron structures are characterized not only by the convergence of information flows but also by their divergence, as well as by making irradiating paths of information spreading. However, the diversity of neuron-detectors responses that determine our subjective perception of external images cannot be explained only by combinatoric character of their interrelation in irradiating paths of signal spreading. It is evident that to create the internal 'world picture', the information that comes only from the receptors of perceptual system is not sufficient. The process of internal presentation of external images (building presentations) should be the process of cognition. The meaning of this process lies in the synthesis of 'knowledge' based on the analysis of information. If the synthesis can be realized by NE-detectors, then to make analysis it is necessary to generate new characteristics that are not formed by the perceptual systems.

Thus, the following hypothesis can be stated.

***Hypothesis 5.*** *Neuronal brain structures perform not only synthesis of neuronal responses but analyze these responses, which is connected with the decomposition of information and formation of new information characteristics of the perceived images. The analysis of information should be conducted by the neurons of other type that functionally and morphologically differ from the neurons-detectors.*

*We call these analytical neurons **NE-analyzers.***

The information models of NE-analyzers will be considered in the coming paper.